\documentclass[11pt,a4paper]{article}
\usepackage[hyperref]{acl2017}
\usepackage{times}
\usepackage{latexsym}
\usepackage{tabularx}
\usepackage{booktabs}
\usepackage{amsmath}
\usepackage{graphicx}
\usepackage{multirow}
\usepackage{tipa}
\usepackage{scrextend}
\usepackage{microtype}

\aclfinalcopy 

\title{Aligning Script Events with Narrative Texts}

\author{Simon Ostermann$^\dagger$ \quad Michael Roth$^{\dagger\ddagger}$ \quad Stefan Thater$^\dagger$ \quad Manfred Pinkal$^\dagger$ \\
  $^\dagger$ Saarland University \quad $^\ddagger$ University of Illinois at Urbana-Champaign\\
  {\tt \{simono\textpipe mroth\textpipe stth\textpipe pinkal\}@coli.uni-saarland.de}}

\date{\today}

\begin{document}
	
\maketitle
\begin{abstract}
Script knowledge plays a central role in text understanding and is relevant for a variety of downstream tasks. 
In this paper, we consider two recent datasets which
provide a rich and general representation of script events in terms of paraphrase sets. 
We introduce the task of mapping event mentions
in narrative texts to such script event types, and present a model for this task that exploits rich
linguistic representations as well as information on temporal ordering. The results of our experiments demonstrate that this complex task is indeed feasible.
\end{abstract}

\section{Introduction}
Event structure is a prominent topic in NLP. While semantic role labelers \cite{Gildea2002,Palmer2010} are well-established tools for the analysis of the internal structure of event descriptions, modeling relations between events has gained increasing attention in recent years. Research on event coreference  
\cite{Bejan2010,Lee2012}, temporal event ordering in newswire texts \cite{Ling2010}, as well as shared tasks on cross-document event ordering \cite[inter alia]{Minard2015} have in common that they model cross-document relations. 

The focus of this paper is on the task of analyzing text-internal event structure. We share the view of a long tradition in NLP (see e.g.~\newcite{Schank1975, Chambers2009, Regneri2010}) that \textit{script knowledge} is of central importance to this task, i.e.~common-sense knowledge about events and their typical order in everyday activities (also referred to as \textit{scenarios}, \newcite{Barr1981}). Script knowledge guides expectation by predicting which type of event or discourse referent might be addressed next in a story \cite{Modi2016b}, allows to infer missing events from events explicitly mentioned \cite{Chambers2009,Jans2012,Rudinger2015b}, and to determine text-internal temporal order \cite{Modi2014,Frermann2014}.

We address the task of automatically mapping narrative texts to scripts, which will leverage explicit script knowledge for the afore-mentioned aspects of text understanding, as well as for downstream tasks such as textual entailment, question answering or paraphrase detection. 

We build on the work of \newcite{Regneri2010} and \newcite{Wanzare2016}, who collect explicit script knowledge via crowdsourcing, by asking people to describe everyday activities. 
These crowdsourced descriptions form a basis for high-quality automatic extraction of script structure without any human intervention \cite{Regneri2010, Wanzare2017}. The events of the resulting structure are defined as sets of alternative realizations, which cover lexical variation and provide paraphrase information. To the best of our knowledge, these advantages have not been explicitly used elsewhere.

Aligning script structures with texts is a complex task. In a first attempt, we assume that three steps are necessary to solve it, although in the long run, an integrated approach will be preferable: First, the script which is addressed by the event mention must be identified. Second, it has to be decided whether a verb denotes a script event at all. 
Finally, event verbs need to be assigned a script-specific event type label. This work focuses on the last two steps: We use a corpus of narrative stories each of which is centered around a specific script scenario, and distinguish verbs related to the central script from all other verb occurrences with a simple decision tree classifier. We then train a sequence labeling model only on crowdsourced script data and assign event type labels to all script-related event verbs.

Our results substantially outperform informed baselines, in spite of the availability of only small amounts of training data. In particular, we also demonstrate the relevance of event ordering information provided by script knowledge. 

Our code and all data and parameters that are used are publicly available under \href{https://github.com/SimonOst}{\texttt{https://github.com/SimonOst}}.

\section{Task and Data}
\label{sec:datatask}

\begin{figure}
	\includegraphics[width=\columnwidth]{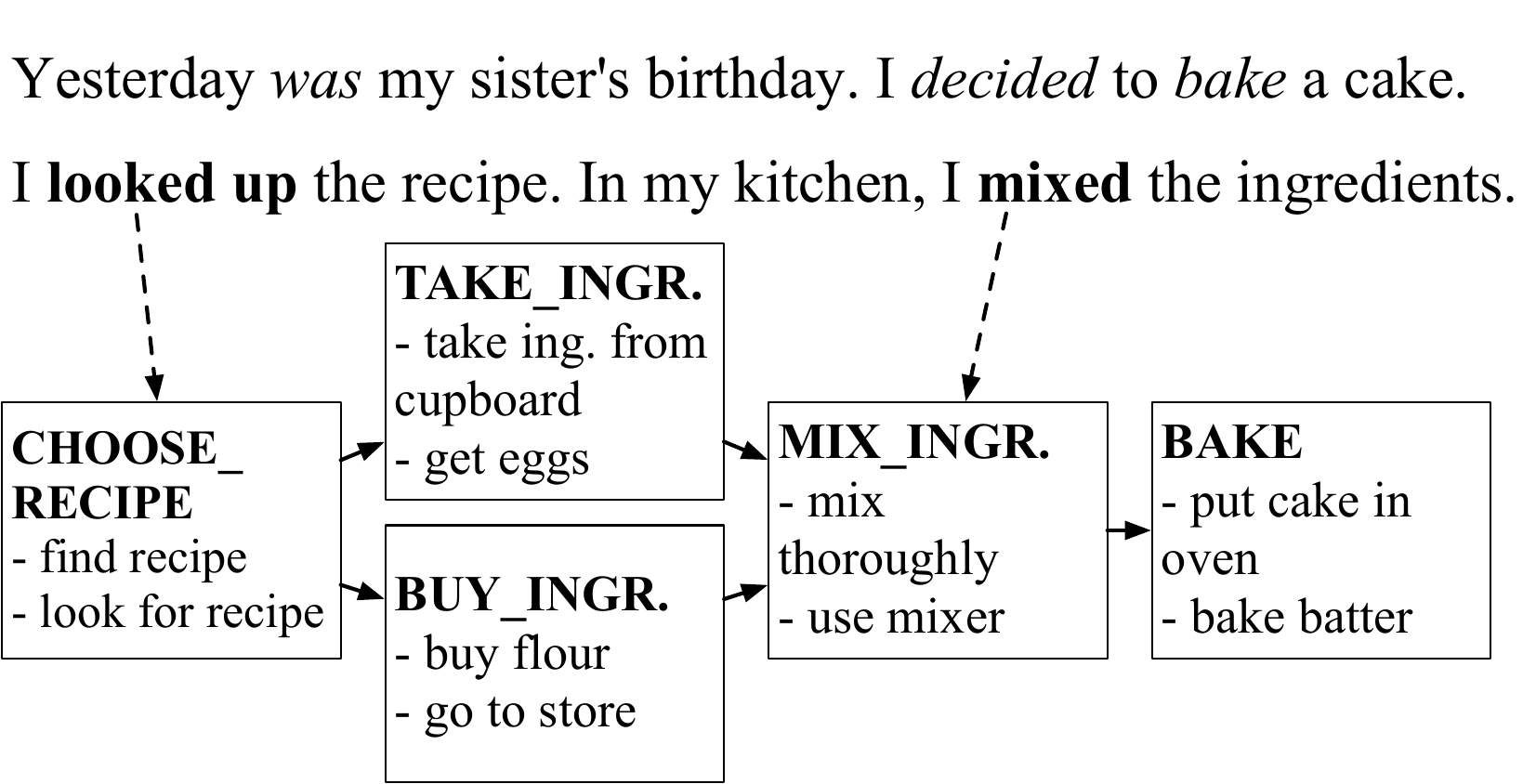}
	\caption{An example of text-to-script mapping with an excerpt of the \textsc{baking a cake} script and a story snippet.}
	\label{fig:ex}
\end{figure}

As a basis for the task of text-to-script mapping, we make use of two recently published datasets. \textit{DeScript} \cite{Wanzare2016} is a collection of crowdsourced linguistic descriptions of event patterns for everyday activities, so called \textit{event sequence descriptions (ESDs)}. ESDs consist of short telegram-style descriptions of single events (\textit{event descriptions, ED}). The textual order of EDs corresponds to the temporal order of respective events, i.e.~temporal information is explicitly encoded. \textit{DeScript} contains 50 ESDs for each of 40 different scenarios. Alongside the ESDs, it also provides gold event paraphrase sets, i.e.~clusters of all event descriptions denoting the same event type, labeled with the respective type. 

While DeScript is a source of structured script knowledge, the \textit{InScript} corpus \cite{Modi2016} provides us with the appropriate kind of narrative texts. \textit{InScript} is a collection of 910 stories centered around some specific scenario, for 10 of the 40 scenarios in \textit{DeScript}, e.g.~\textsc{baking a cake}, \textsc{riding a bus}, \textsc{taking a shower}. All verbs occurring in the texts are annotated with an event type if they are relevant to the script instantiated by the story; as \textit{non-script event} otherwise.

In the upper part of Fig.~\ref{fig:ex}, you see the initial fragment of a story about baking a cake; together with a script excerpt in the lower part, depicted by labeled event paraphrase sets. \textit{I looked up the recipe} and \textit{I mixed the ingredients} mention relevant script events, and therefore should be labeled with the indicated event types (\textsc{choose\_recipe}, \textsc{mix\_ingredients}). Fig.~\ref{fig:ex} also illustrates the potential of text-to-script mapping: script knowledge  enables to predict that a baking event might be addressed  next in the story. The verb \textit{was} does not denote an event at all, and \textit{decide} is not part of the \textsc{baking a cake} script, so they are assigned the label \textit{non-script event}. Actually, \textit{InScript} comes with two additional categories of verbs (\textit{script-related} and \textit{script-evoking}), which we subsume under \textit{non-script event}.

The central task addressed in our paper, the automatic labeling of all script-relevant verbs in the \textit{InScript} text with a script-specific event type, uses only \textit{DeScript} data for training; event-type labels of \textit{InScript} are used for evaluation purposes only.

\section{Model}

Section~\ref{sec:event-type-identification} defines the central part of our system, a sequence model for classifying script-relevant verbs into scenario-specific event types. For full automation of the text-to-script mapping, we describe in Section~\ref{sec:event-identification} a model for identifying script-relevant verbs.

\subsection{Event Type Classification}
\label{sec:event-type-identification}
For identifying the correct event type given a script-relevant verb, we leverage two types of information: We require a representation for the meaning and content of the event mention, which takes into account not only the verb, but also the persons and objects involved in an event, i.e. the \textit{script participants}. In addition, we take event ordering information into account, which helps to disambiguate event mentions based on their local context. To model both event types and sequences thereof, we implement a linear-chain conditional random field (CRF, \newcite{Lafferty2001}). Our implementation is based on the CRF++ toolkit\footnote{\href{https://taku910.github.io/crfpp/}{\texttt{taku910.github.io/crfpp/}}} and employs two types of features:

\textbf{Sequential Feature.} Our CRF model utilizes event ordering information in the form of binary indicator features that encode the co-occurrence of two event type labels in sequence.

\textbf{Meaning Representation Features.} Two feature types encode the meaning of a textual event mention. One is a shallow form of representation derived from precomputed word embeddings (\textit{word2vec}, \newcite{Mikolov2013}). This feature type captures distributional information of the verb and its direct nominal dependents\footnote{For EDs, we use all mentioned head nouns.}, which we assume to denote script participants, and is computed by averaging over the respective word vector representations.\footnote{To emphasize the importance of the verb, we double its weight when averaging.} We use pretrained 300-dimensional embeddings that are trained on the Google News corpus.\footnote{Because our CRF model only supports nominal features, we discretize embeddings from \href{https://code.google.com/archive/p/word2vec/}{\texttt{code.google.com/archive/p/word2vec/}} by binning the component values into three intervals $[-\infty,-\epsilon], [-\epsilon,\epsilon], [\epsilon,\infty]$. The hyperparameter $\epsilon$ is determined on a held-out development set.}
As a more explicit but sparse form of content representation, we use as the other type of feature the lemma of the verb, its indirect object and its direct object.

\subsection{Identifying Script-Relevant Verbs}
\label{sec:event-identification}
We use a decision tree classifier for identifying script-relevant verbs (\textit{J48} from the Weka toolkit, \newcite{Frank2016}) that takes into account four classes: the three \textit{non-script event} classes from \textit{InScript} and one class for all \textit{event-verbs}. At test time, the three \textit{non-script event} classes are merged into one class. Due to the lack of \textit{non-script event} instances in \textit{DeScript}, we train and test our model on all verbs occurring in \textit{InScript}. We use the following feature types: 

\textbf{Syntactic Features.} We employ syntactic features for identifying verbs that only rarely denote script events, independent of the scenario:
a feature for auxiliaries; for verbs that govern an adverbial phrase (mostly if-clauses); a feature indicating the number of direct and indirect objects; and a lexical feature that checks if the verb belongs to a predefined list of non-action verbs.

\textbf{Script Features.} For finding verbs that match the current script scenario, we employ two features: a binary feature indicating whether the verb is used in the ESDs for the given scenario; and a scenario-specific tf--idf score that is computed by treating all ESDs from a scenario as one document, summed over the verb and its dependents. In Section~\ref{subsec:eval-identification}, we evaluate models with and without script features, to test the impact of scenario-specific information.

\textbf{Frame Feature.} We further employ frame-semantic information because we expect script events to typically evoke certain frames. 
We use a state-of-the-art semantic role labeler \cite{Roth2016a,Roth2016b} based on \textit{FrameNet} \cite{Ruppenhofer2006} to predict frames for all verbs, encoding the frame as a feature. We address sparsity of too specific frames by mapping all frames to higher-level super frames using the \textit{framenet querying package}\footnote{\href{https://www.github.com/icsi-berkeley/framenet}{\texttt{github.com/icsi-berkeley/framenet}}}.

\begin{table}
	\centering
	\begin{tabularx}{\columnwidth}{lXXX}
		\toprule
		& \textbf{~~P} & \textbf{~~R} & \textbf{~~F$_1$}\\
		\midrule
		\textit{Lemma}  &	0.365 & \textbf{0.949} & 0.526\\
		\textit{Our model} & \textbf{0.628} & 0.817 & \textbf{0.709} \\
		\textit{Our model (scen. indep.)} & 0.513 &	0.877	& 0.645 \\
		\bottomrule
	\end{tabularx}
	\caption{Identification of script-relevant verbs within a scenario and independent of the scenario.}
	\label{tab:eval:ei}
\end{table}

\section{Evaluation}

\subsection{Experimental Setup}

We evaluate our model for text-to-script mapping based on the resources introduced in Section~\ref{sec:datatask}. We process the \textit{InScript} and \textit{DeScript} data sets using the Stanford Parser \cite{Klein2003}\footnote{To improve performance on the simplistic sentences from \textit{DeScript}, we follow \newcite{Regneri2013} and re-train the parser.}. We further resolve pronouns in \textit{InScript} using annotated coreference chains from the gold standard.

We individually test the two components, i.e. the identification of script-relevant verbs and event classification. Experiments on the first sub-task are described in Section~\ref{subsec:eval-identification}. Sections~\ref{subsec:eval-classification} and~\ref{subsec:eval-joint} present results on the latter task and a combination of both tasks, respectively.

\subsection{Identifying Script-Relevant Verbs}
\label{subsec:eval-identification}
In this evaluation, we test the ability of our model to identify verbs in narrative texts that instantiate script events. Our experiments make use of a 10-fold cross-validation setting within all texts of one scenario. To test the model in a scenario-independent setting, we perform additional experiments based on a cross-validation with the 10 scenarios as one fold each and exclude the script features. That is, we repeatedly train our model on 9 scenarios and evaluate on the remaining scenario, without using any information about the test scenario. 

\textbf{Models.} We compare the model described in Section~\ref{sec:event-identification} to a baseline (\textit{Lemma}) that always assigns the \textit{event} class if the verb lemma is mentioned in \textit{DeScript}. We report precision, recall and F$_1$-score on event verbs, averaged over all scenarios. 

\textbf{Results.} Table \ref{tab:eval:ei} gives an overview of the results based on 10-fold cross-validation. Our scenario-specific model is capable of identifying more than 81\% of script-relevant verbs at a precision of about 63\%. This is a notable improvement over the baseline, which identifies 94.9\% of the event verbs, but at a precision of only 36.5\%.

The table also gives numbers for the scenario-independent setting: Precision drops to around 51\% if only training data from other scenarios is available. One of the main difficulties here lies in classifying different \textit{non-script event} verb classes in a way that generalizes across scenarios. \newcite{Modi2016} also found that distinguishing specific types of non-script events from script events can be difficult even for humans.

\subsection{Event Type Classification}
\label{subsec:eval-classification}
In this section, we describe experiments on the text-to-script mapping task based on the subset of event instances from \textit{InScript} that are annotated as script-related. As training data, we use the \textit{ESDs} and the event type annotations from the \textit{DeScript} gold standard\footnote{In \textit{DeScript}, there are some rare cases of \textit{EDs} that do not describe a script event, but that are labeled as \textit{non-script event}. We exclude these from the training data.}. 
The evaluation task is to classify individual event mentions in \textit{InScript} based on their verbal realization in the narrative text. We evaluate against the gold-standard annotations from \textit{InScript}. Since event type annotations are used for evaluation purposes only, this task comes close to a realistic setup, in which script knowledge is available for specific scenarios but no training data in the form of event-type annotated narrative texts exists.

\textbf{Models.} We evaluate our CRF model described in Section~\ref{sec:event-type-identification} against two baselines that are based on textual similarity. Both baselines compare the event verb and its dependents in \textit{InScript} to all EDs in \textit{DeScript} and assign the event type with the highest similarity. \textit{Lemma} is a simple measure based on word overlap, \textit{word2vec} uses the same embedding representation as the CRF model (before discretization) but simply assigns the best matching event type label based on cosine similarity. 
We report precision, recall and F$_1$-scores, macro-averaged over all script-event types and scenarios.

\begin{table}
	\centering
	\begin{tabularx}{\columnwidth}{lXXX}
		\toprule
		~~~~~~~~~~~~~~~~~~~~ & \textbf{~~P} & \textbf{~~R} & \textbf{~~F$_1$} \\
		\midrule
		\textit{Lemma}        & 0.516 & 0.442 & 0.475 \\
		\textit{Word2Vec}     & 0.538 & 0.480 & 0.507 \\
		\textit{CRF model}    & \textbf{0.623} & \textbf{0.485 }& \textbf{0.543} \\
		\textit{CRF, no seq.} & 0.608 & 0.475 & 0.531\\
		\bottomrule
	\end{tabularx}
	\caption{Event Type Classification performance, with and without sequential features.}
	\label{tab:eval:gs}
\end{table}

\textbf{Results.} Results for all models are presented in Table~\ref{tab:eval:gs}.\footnote{\label{note}Updated in 04/19 due to a bug in the evaluation.} Our CRF model achieves a F$_1$-score of 0.543, a considerably higher performance in comparison to the baselines. As can be seen from excluding the sequential feature, ordering information improves the result. The rather small difference is due to the fact that ordering information can also be misleading (cf.~Section \ref{sec:discussion}). We found, however, that including the sequential feature accounts for an improvement of up to 4\% in F$_1$ score, depending on the scenario.

\begin{table}
	\centering
	\begin{tabularx}{\columnwidth}{lXXX}
		\toprule
		~~~~~~~~~~~~~~~~~~~~ & \textbf{~~P} & \textbf{~~R} & \textbf{~~F$_1$} \\
		\midrule
		\textit{Ident.\ model}$+$\textit{Lemma}     & 0.388 & 0.475 & 0.426\\
		\textit{Ident.\ model}$+$\textit{Word2vec}  &0.393 & \textbf{0.511} &0.442 \\
		\textit{Ident.\ model}$+$\textit{CRF model} & \textbf{0.458} &0.505 & \textbf{0.478} \\
		\bottomrule
	\end{tabularx}
	\caption{Full text-to-script mapping results.}
	\label{tab:eval:eti}
\end{table}

\subsection{Full Text-to-Script Mapping Task}
\label{subsec:eval-joint}
We now address the full text-to-script mapping task, a combination of the identification of relevant verbs and event type classification. This setup allows us to assess whether the general task of a fully automatic mapping of verbs in narrative texts to script events is feasible.

\textbf{Models.} We compare the same models as in Section~\ref{subsec:eval-classification}, but use them on top of our model for identifying script-relevant verbs (cf. Section \ref{subsec:eval-identification}) instead of using the gold standard for identification.

\textbf{Results.} On the full text-to-script mapping task, our combined identification and CRF model achieves a precision and recall of $0.458$ and $0.505$, resp.\ (cf.\ Table~\ref{tab:eval:eti}).\footref{note} This reflects an absolute improvement over the baselines of $0.036$ and $0.052$ in terms of F$_1$-score. The results reflect the general difficulty of this task but are promising overall. As reported by \newcite{Modi2016}, even human annotators only achieve an agreement of 0.64 in terms of Fleiss' Kappa (1971\nocite{Fleiss1971}).

\section{Discussion}
\label{sec:discussion}
In this section, we discuss cases in which our system predicted the wrong event type and give examples for each case. We found 3 major error sources:

\textbf{Lexical Coverage.} We found that although \textit{DeScript} is a small resource, training a model purely on \textit{ESDs} works reasonably well. Coverage problems can be seen in cases of events for which only few \textit{EDs} exist. An example is the \textsc{choose\_tree} event (the event of picking a tree at the shop) in the \textsc{planting a tree} scenario. There are only 3 \textit{EDs} describing the event, each of which uses the event verb ``choose''. In contrast, we find that ``choose'' is used in less than 10\% of the event mentions in \textit{InScript}. Because of this mismatch, which can be attributed to the small training data size, more frequently used verbs for this event in \textit{InScript}, such as ``pick'' and ``decide'', are labeled incorrectly.

We observe that our meaning representation might be insufficient for finding synonyms for about 30\% of observed verb tokens. This specifically includes scenario-specific and uncommon verbs, such as ``squirt'' in the context of the \textsc{baking a cake} scenario (\textsl{squirt the frosting onto the cake}). Problems may also arise from the fact that about 23\% of the verb types occur in multiple paraphrase clusters of a scenario.

\textbf{Misleading Ordering Information.} We found that ordering information is in general beneficial for text-to-script alignment. We however also identified cases for which it can be misleading, by comparing the output of our full model to the model that does not use sequential features. As another result of the small size of \textit{DeScript}, there are plausible event sequences that appear only rarely or never in the training data. This error source is involved in 60--70\% of the observed misclassifications due to misleading ordering information. An example is the \textsc{wash} event in the \textsc{getting a haircut} scenario: It never appears directly after the \textsc{move\_in\_salon} event (i.e.~walking from the counter to the chair) in \textit{DeScript}, but its a plausible sequence that is misclassified by our model.

In almost 15\% of the observed errors, an event type is mentioned more than once, leading to misclassifications whenever ordering information is used. One reason for this might be that events in \textit{InScript} are described in a more exhaustive or fine-grained way. For example, the \textsc{wash} event in the \textsc{taking a bath} scenario is often broken up into three mentions: wetting the hair, applying shampoo, and washing it again. However, because there is only one event type for the three mentions, this sequence is never observed in \textit{DeScript}. 

Events with an interchangeable natural order lead to errors in a number of cases: In the \textsc{baking a cake} scenario, a few misclassifications happen because the order in which e.g. ingredients are prepared, the pan is greased and the oven is preheated is very flexible, but the model overfits to what it observed from the training.

As last, there are also a few cases in which an event is mentioned, even before it actually takes place. In the case of the \textit{borrowing a book} scenario, there are cases in \textit{InScript} that mention in the first sentence that the purpose of the visit is to return a book. In \textit{DeScript} in contrast, the \textsc{return} event always takes place in the very end.

\textbf{Near Misses.} For many verbs, it is also difficult for humans to come up with one correct event label. By investigating confusion matrices for single scenarios, we found that for at least 3--5\% of script event verbs in the test set, our model predicted an ``incorrect'' label for such verbs, but that label might still be plausible. In the \textsc{baking a cake} scenario, for example, there is little to no difference between mentions of making the dough and preparing ingredients. As a consequence, these two events are often confused: Approximately 50\% of the instances labeled as \textsc{prepare\_ingredients} are actually instances of \textsc{make\_dough}. 

\section{Summary}
In this paper, we addressed the task of automatically mapping event denoting expressions in narrative texts to script events, based on an explicit script representation that is learned from crowdsourced data rather than from text collections. Our models outperform two similarity-based baselines by leveraging rich event representations and ordering information. We showed that models of script knowledge can be successfully trained on crowdsourced data, even if the number of training examples is small. This work thus builds a basis for utilizing the advantages of crowdsourced script representations for downstream tasks and future work, e.g. paraphrase identification in discourse context or event prediction on narrative texts. 

\section*{Acknowledgments}
We thank the anonymous reviewers for their helpful comments. This research was funded by the German Research Foundation (DFG) as part of SFB~1102 `Information Density and Linguistic Encoding'. Work by MR in Illinois was supported by a DFG Research Fellowship (RO 4848/1-1).

\bibliography{references}
\bibliographystyle{acl_natbib}

\end{document}